\documentclass{article}

\PassOptionsToPackage{numbers, compress}{natbib}
\usepackage[final]{nips_2017}

\usepackage[utf8]{inputenc} 
\usepackage[T1]{fontenc}    
\usepackage{hyperref}       
\usepackage{url}            
\usepackage{bm}
\usepackage{booktabs}       
\usepackage{amsfonts}       
\usepackage{amsmath}
\usepackage{nicefrac}       
\usepackage{microtype}      
\usepackage{algorithm,algpseudocode}
\usepackage{graphicx}
\usepackage{listings}
\usepackage{tikz}
\usepackage{xcolor}
\usepackage{sidecap}
\usepackage{siunitx}

\newcommand{\RR}{\mathbb{R}}
\DeclareMathOperator*{\argmax}{arg\,max}

\newcommand{\xx}{{\mathbf{x}}}

\definecolor{maroon}{cmyk}{0, 0.87, 0.68, 0.32}
\definecolor{halfgray}{gray}{0.55}
\definecolor{ipython_frame}{RGB}{207, 207, 207}
\definecolor{ipython_bg}{RGB}{247, 247, 247}
\definecolor{ipython_red}{RGB}{186, 33, 33}
\definecolor{ipython_green}{RGB}{0, 128, 0}
\definecolor{ipython_cyan}{RGB}{64, 128, 128}
\definecolor{ipython_purple}{RGB}{170, 34, 255}

\usetikzlibrary{bayesnet}

\lstdefinelanguage{iPython}{
    morekeywords={access,and,break,class,continue,def,del,elif,else,except,exec,finally,for,from,global,if,import,in,is,lambda,not,or,pass,print,raise,return,try,while},%
    morekeywords=[2]{abs,all,any,basestring,bin,bool,bytearray,callable,chr,classmethod,cmp,compile,complex,delattr,dict,dir,divmod,enumerate,eval,execfile,file,filter,float,format,frozenset,getattr,globals,hasattr,hash,help,hex,id,input,int,isinstance,issubclass,iter,len,list,locals,long,map,max,memoryview,min,next,object,oct,open,ord,pow,property,range,raw_input,reduce,reload,repr,reversed,round,set,setattr,slice,sorted,staticmethod,str,sum,super,tuple,type,unichr,unicode,vars,xrange,zip,apply,buffer,coerce,intern},%
    sensitive=true,%
    morecomment=[l]\#,%
    morestring=[b]',%
    morestring=[b]",%
    morestring=[s]{'''}{'''},
    morestring=[s]{"""}{"""},
    morestring=[s]{r'}{'},
    morestring=[s]{r"}{"},%
    morestring=[s]{r'''}{'''},%
    morestring=[s]{r"""}{"""},%
    morestring=[s]{u'}{'},
    morestring=[s]{u"}{"},%
    morestring=[s]{u'''}{'''},%
    morestring=[s]{u"""}{"""},%
    literate=
    {á}{{\'a}}1 {é}{{\'e}}1 {í}{{\'i}}1 {ó}{{\'o}}1 {ú}{{\'u}}1
    {Á}{{\'A}}1 {É}{{\'E}}1 {Í}{{\'I}}1 {Ó}{{\'O}}1 {Ú}{{\'U}}1
    {à}{{\`a}}1 {è}{{\`e}}1 {ì}{{\`i}}1 {ò}{{\`o}}1 {ù}{{\`u}}1
    {À}{{\`A}}1 {È}{{\'E}}1 {Ì}{{\`I}}1 {Ò}{{\`O}}1 {Ù}{{\`U}}1
    {ä}{{\"a}}1 {ë}{{\"e}}1 {ï}{{\"i}}1 {ö}{{\"o}}1 {ü}{{\"u}}1
    {Ä}{{\"A}}1 {Ë}{{\"E}}1 {Ï}{{\"I}}1 {Ö}{{\"O}}1 {Ü}{{\"U}}1
    {â}{{\^a}}1 {ê}{{\^e}}1 {î}{{\^i}}1 {ô}{{\^o}}1 {û}{{\^u}}1
    {Â}{{\^A}}1 {Ê}{{\^E}}1 {Î}{{\^I}}1 {Ô}{{\^O}}1 {Û}{{\^U}}1
    {œ}{{\oe}}1 {Œ}{{\OE}}1 {æ}{{\ae}}1 {Æ}{{\AE}}1 {ß}{{\ss}}1
    {ç}{{\c c}}1 {Ç}{{\c C}}1 {ø}{{\o}}1 {å}{{\r a}}1 {Å}{{\r A}}1
    {€}{{\EUR}}1 {£}{{\pounds}}1
    {^}{{{\color{ipython_purple}\^{}}}}1
    {=}{{{\color{ipython_purple}=}}}1
    {+}{{{\color{ipython_purple}+}}}1
    {*}{{{\color{ipython_purple}$^\ast$}}}1
    {/}{{{\color{ipython_purple}/}}}1
    {+=}{{{+=}}}1
    {-=}{{{-=}}}1
    {*=}{{{$^\ast$=}}}1
    {/=}{{{/=}}}1,
    literate=
    *{-}{{{\color{ipython_purple}-}}}1
     {?}{{{\color{ipython_purple}?}}}1,
    identifierstyle=\color{black}\ttfamily,
    commentstyle=\color{ipython_cyan}\ttfamily,
    stringstyle=\color{ipython_red}\ttfamily,
    keepspaces=true,
    showspaces=false,
    showstringspaces=false,
    rulecolor=\color{ipython_frame},
    frame=single,
    frameround={t}{t}{t}{t},
    framexleftmargin=6mm,
    framexrightmargin=0mm,
    numbers=left,
    numberstyle=\tiny\color{halfgray},
    backgroundcolor=\color{ipython_bg},
    basicstyle=\scriptsize,
    keywordstyle=\color{ipython_green}\ttfamily,
}

\title{Sequential Preference-Based Optimization}

\author{
  Ian Dewancker \ \ \ \ \ \ Jakob Bauer \ \ \ \\
  Uber Advanced Technologies Group\\
  Pittsburgh, PA  \\
  \texttt{\{idewancker, jbauer1\}@uber.com}
  \And
  Michael McCourt \\
  SigOpt\\
  San Francisco, CA  \\
  \texttt{mccourt@sigopt.com} \\
}

\begin{document}

\maketitle

\begin{abstract}
  Many real-world engineering problems rely on human preferences to guide their design and optimization.  
  We present PrefOpt, an open source package to simplify sequential optimization tasks that incorporate human preference feedback.  Our approach extends an existing latent variable 
  model for binary preferences to allow for observations of 
  equivalent preference from users.
\end{abstract}

\section{Introduction}

Defining metrics amenable to optimization can be quite challenging in complex engineering systems.  Systems often have components that
rely on human perception to judge performance and it can be difficult to develop quantitative measures to capture these perceptual metrics.
We are motivated by the problem of tuning the behavior of a motion planning system for a driverless car.
In this setting, it may be more practical to employ a tuning strategy that only requires a user to conduct pairwise comparisons of planning system configurations as either more, less or equivalently 
comfortable.
\begin{figure}[H]
 \begin{center}
 \includegraphics[width=0.5\textwidth]{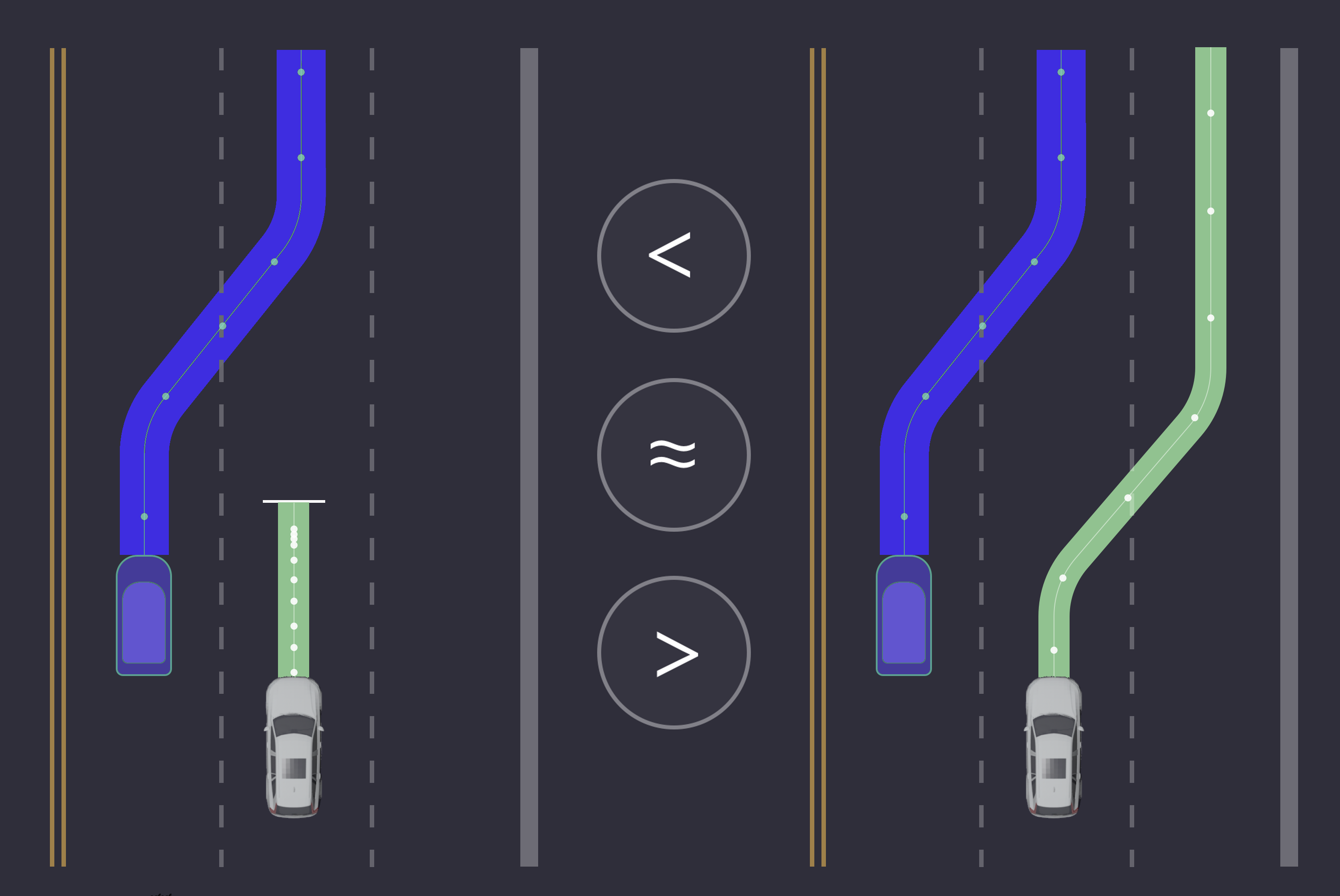} 
 \end{center}
 \caption{Hypothetical preference-based optimization of a motion planning system.  A user is asked to compare 
 a sequence of configurations of a planning system as either worse, better or equivalent.  The pairwise comparisons are used to refine the search for the optimal configuration.}
 \label{fig:pref_compare}
\end{figure}

\vspace{-2mm}
By only requiring comparative judgements as feedback, the user is freed from developing a rigorous, single scalar measurement of comfort. 
Human-in-the-loop preference-based optimization methods
\cite{brochu2010bayesian,eric2008active,houlsby2012collaborative, dewancker2016,gonzalez2017preferential,thatte2017sample} have 
been developed to alleviate some of 
the requirements of previous model-based optimization methods \cite{bergstra2011algorithms,hutter2011sequential,snoek2012practical}.  These interactive optimization
methods allow users to more 
easily optimize systems that are measured using perceptual metrics or multiple metrics.  In this article, we introduce an extension to an existing preference model, 
allowing for the user to report configurations as equivalently preferable.  We also introduce PrefOpt, an open source library that builds on Edward \cite{tran2016edward} to conduct sequential
preference-based optimization using the extended preference model.

%

\section{Preference Model Supporting Ties}
\label{gen_inst}

Models that relate discrete preference observations to latent function values drawn from Gaussian process priors have been well studied \cite{chu2005preference,eric2008active,guo2010gaussian,brochu2010bayesian}.
Previous preference models have required that the user state a binary preference when presented with
two options ($\xx_i,\xx_j\in\Omega\subset\RR^D$). In many real-world applications, it may be that even 
experts occasionally have difficulty discerning two alternatives in terms of absolute preference.  To address this concern, we extend the discrete preference observations with a third option:
specifying equivalent preference between the two alternatives. 
Specifically, we adopt a modified Bradley-Terry model that supports ties, or configurations with equivalent preference \cite{rao1967ties}. 
\vspace{-4mm}
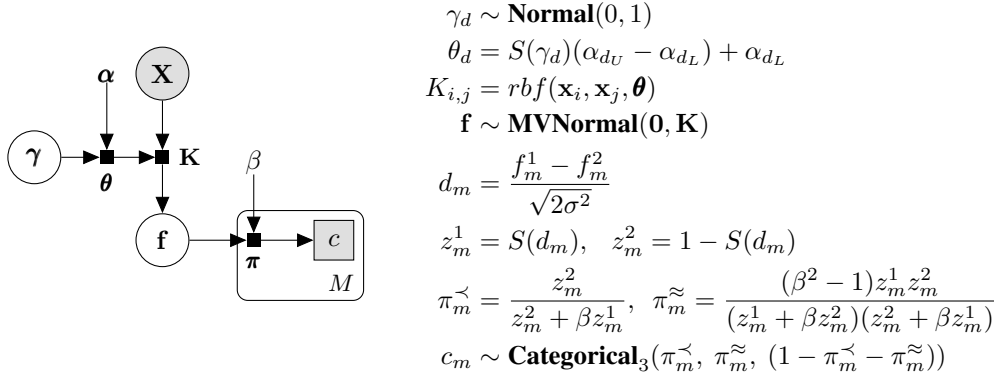
\begin{figure}[h]
	\centering
	\begin{minipage}{.4\textwidth}
		\centering
		\begin{tikzpicture}
		
		  \node[obs_disc]		          (c)     {$c$};
		  \node[latent, left=1.65cm of c]         (f)     {$\mathbf{f}$};
		  \node[obs, above=1.5cm of f]            (X)     {$\mathbf{X}$};
		  \factor[above=0.65cm of f] {K} {right:$\mathbf{K}$} {}{}; %
		
		  \factor[left=0.7cm of c]          {z}     {below:$\pmb{\pi}$} {}{};
		
		  \factor[left=0.55cm of K] {theta} {below:$\pmb{\theta}$} {}{};
		  \node[latent, left=1.25cm of K]         (gamma) {$\pmb{\gamma}$};
		  \node[const, above=0.78cm of z]        (beta)  {$\beta$};
		  \node[const, above=0.90cm of theta]      (alpha) {$\pmb{\alpha}$};
		  \edge [] {z} {c} ; %
		  \edge {f} {z} ; %
		  \edge {theta} {K} ; %
		  \edge {gamma} {theta} ; %
		  \edge {X} {K} ; %
		  \edge {K} {f} ; %
		  \edge {beta}  {z} ; %
		  \edge {alpha} {theta} ; %
		
		  \plate {zc} {(z)(c)} {$M$} ;
		  
		\end{tikzpicture}
	\end{minipage}%
	\begin{minipage}{.6\textwidth}
		\centering
		\begin{align*}
			\gamma_d &\sim \textbf{Normal}(0, 1)   \\
			\theta_d &= S(\gamma_d)(\alpha_{d_U} - \alpha_{d_L}) + \alpha_{d_L} \\
			K_{i,j} &= rbf(\mathbf{x}_{i}, \mathbf{x}	_{j}, \pmb{\theta}	) \\
			\mathbf{f} &\sim \textbf{MVNormal}(\mathbf{0}, \mathbf{K}) \\
			d_m &= \frac{f_{m}^1 - f_{m}^2}{\sqrt{2\sigma^2}} \\
			z^{1}_m &= S(d_m), \ \ \ z^{2}_m = 1 - S(d_m)\ \\
			\pi^{\prec}_m &= \frac{z^2_m}{z^2_m+\beta z^1_m}, \ \ \pi^{\approx}_m = \frac{(\beta^2-1)z^1_mz^2_m}{(z^1_m+\beta z^2_m)(z^2_m+\beta z^1_m)} \\
			c_m &\sim \textbf{Categorical}_3(\pi^{\prec}_m, \ \pi^{\approx}_m, \ (1 - \pi^{\prec}_m - \pi^{\approx}_m))
		\end{align*}
		\end{minipage}
	\caption{
		\label{fig:graphicalmodel}
		Graphical model \emph{(left)} and generative process \emph{(right)} of the preference model.
	}
\end{figure}

The model draws latent function vectors $\mathbf{f} \in \RR^N$ from a Gaussian process prior where each entry
corresponds to one of the $N$ unique query points $(\mathbf{X})$ the user has compared so far.  The $\pmb{\gamma}$ variables are drawn from normal priors and then transformed to form the length-scales $\pmb{\theta}$ of the 
covariance function; they, in turn, are used to define $\mathbf{K}$, the covariance matrix.
Length scales are bounded by $\pmb{\alpha}$ which, along with $S(x) = \frac{1}{1 + e^{-x}}$,
is used to define the transformation from $\pmb{\gamma}$ to $\pmb{\theta}$.
Here, $rbf(\mathbf{x}_i,\mathbf{x}_j,\pmb{\theta}) = \sigma^2 \exp\Big(-\frac{1}{2} \sum_{d=1}^D \frac{1}{\theta_d^2} ({x_i}_d - {x_j}_d)^2 \Big)$.

The generalized Bradley-Terry model relates the observed discrete preference data $c_m$ to the latent function values $(f_m^1, f_m^2)$ associated with the two points $(\mathbf{x}_m^1, \mathbf{x}_m^2)$ compared 
by the user during an interactive query.  The tie parameter $\beta \geq 1$, is inversely related to the precision
with which a user can state a preference \cite{rao1967ties}.  A higher value for $\beta$ leads to more mass being place in the equivalence bin $(\pi^{\approx}_m)$ of the categorical distribution over the three
possible preference outcomes for two query points.

\subsection{Variational Inference}

In place of approximating the posterior with a multivariate Gaussian using the Laplace approximation around a MAP estimate of the latent variables \cite{chu2005preference,guo2010gaussian}, we 
opt for an approximation that employs variational inference.  We set out to 
approximate $p( \mathbf{z} \ | \ \mathbf{X}, \mathbf{c})$, the posterior of the latent random variables, where $\mathbf{z} = \{\mathbf{f}, \pmb{\gamma}\}$ is the combined set of latent random variables in our model.  
We use a mean field approximation strategy to construct our approximating distribution $q$ : a factored set of Gaussians each parametrized by a mean and variance
as shown below.
\begin{align*}
 p( \mathbf{z} \ | \ \mathbf{X}, \mathbf{c}) \approx q(\mathbf{z} \ ; \ \pmb{\lambda} ) &= \prod_{i=1}^{N} \mathcal{N}(z_i \ | \ {\lambda_{\mu}}_i, \ {\lambda_{\sigma}}_i) \prod_{k=1}^{D} \mathcal{N}(z_k \ | \ {\lambda_{\mu}}_k, \ {\lambda_{\sigma}}_k)
\end{align*}
We rely on the techniques built into Edward \cite{tran2016edward} to perform the optimization required to recover the variational parameters $\pmb{\lambda}$ that minimize the reverse KL divergence between
the true posterior distribution $p$ and the approximating distribution $q$.
In total there will be $2N + 2D$ variational parameters; two for each of the $N$ entries in $\mathbf{f}$ and two for each of the $D$ elements of $\pmb{\gamma}$.



\section{Acquisition Function for Preference-Based Optimization}
\label{headings}

To determine the next point ($\mathbf{x}^n$) to be presented to the user as a comparison point, we adopt a strategy that searches the domain for where the expected improvement of the latent function  
is highest relative to the current, most preferred point ($\mathbf{x}^b$) \cite{eric2008active}.  With our approximation $q(\mathbf{z} \ ; \ \pmb{\lambda} )$  
of the posterior, it is possible to explore the use of an integrated acquisition function, as proposed in \cite{snoek2012practical}. 
\vspace{1.5mm}
\begin{align*}
\mathbf{k}_* &= [rbf(\mathbf{x}^*,\mathbf{x}_1, \pmb{\theta}) \ \cdots \ rbf(\mathbf{x}^*,\mathbf{x}_N, \pmb{\theta}) ] \\
\mu(\mathbf{x}^*) &= \mathbf{k}_{*}^{\mathsf{T}} \mathbf{K}^{-1} \mathbf{f} \\
s^2(\mathbf{x}^*) &= rbf(\mathbf{x}^*,\mathbf{x}^*, \pmb{\theta}) - \mathbf{k}_{*}^{\mathsf{T}} \mathbf{K}^{-1}\mathbf{k}_{*} \\
d &= \mu(\mathbf{x}^*) - f_{best} \\
a_{\mathsf{EI}}(\mathbf{x}^* ; \mathbf{z}  ) &= \begin{cases}
d\Phi(\frac{d}{s(\mathbf{x}^*)}) + s(\mathbf{x}^*)\phi(\frac{d}{s(\mathbf{x}^*)})   , \hspace{2mm} \text{if} \ s(\mathbf{x}^*) > 0\\
0 , \hspace{4cm} \text{if} \ s(\mathbf{x}^*) = 0
\end{cases} \\
\mathbf{x}^n &= \argmax_{\mathbf{x}^*} \int a_{\mathsf{EI}}(\mathbf{x}^* ; \mathbf{z}  ) q(\mathbf{z} \ ; \ \pmb{\lambda} ) d\mathbf{z}
\end{align*}

Here $\Phi(\cdot)$ and $\phi(\cdot)$ denote the CDF and PDF of
the standard normal distribution, respectively.  The value $f_{best}$ is the latent function value associated with the currently most preferred configuration $\mathbf{x}^b$.
If the user is always asked to compare $\mathbf{x}^n$ against the current most preferred point $\mathbf{x}^b$, the most preferred point can be updated as a result of this
comparison.

\section{PrefOpt Software}

The goal of the PrefOpt software package is to provide a simple interface for conducting human-in-the-loop, preference-based optimization tasks. 
The user is required to initially define a bounding box to represent the search domain of the parameters of interest.
The optimization proceeds by iteratively proposing two query points for the user to compare \ref{fig:progam_example}.
The preference order of the two query points is recorded and the underlying latent preference model is updated.
Currently only two query points will be returned each iteration for the user to compare, however, a strategy of returning larger batches of comparison points could be pursued 
\cite{brochu2010bayesian, contal2013parallel}. 

\begin{figure}[h]
\begin{center}
\begin{lstlisting}[language=iPython, xleftmargin=6mm]
import prefopt
# define the domain of the search space
bounding_box = [[-5.0, 5.0], [0.0, 10.0]]
exp = prefopt.PreferenceExperiment(bounding_box)

for i in xrange(1,N):
    # search for the next points to compare
    X = exp.find_next()    
    # get user preference : -1 denotes x1 < x2,  0 denotes x1 = x2, 1 denotes x1 > x2 
    order = get_user_pref(X[0], X[1])
    # update model with new preference observation
    exp.prefer(X[0], X[1], order)
\end{lstlisting}
\end{center}
	\caption{
		\label{fig:progam_example}
		Example usage of PrefOpt to conduct preference-based optimization.
	}
\end{figure}

The PrefOpt library will hopefully facilitate future investigations into preference-based optimization and similar interactive optimization techniques.  In particular, it is exciting to consider applications in the 
self-driving space where metrics are often difficult to specify or intrinsically linked to human perception.  Other interesting avenues of future work 
could include investigating strategies to propose multiple comparison query points at each iteration, evaluating new acquisition functions, or comparing the effectiveness of 
non-GP based methods for capturing user preference data.

The open source PrefOpt package will be hosted at \url{https://github.com/prefopt/prefopt}.

\section{Experimental Results}
\begin{figure}[ht!] \label{fig7} 
  \begin{minipage}[b]{0.5\linewidth}
    \includegraphics[width=\linewidth]{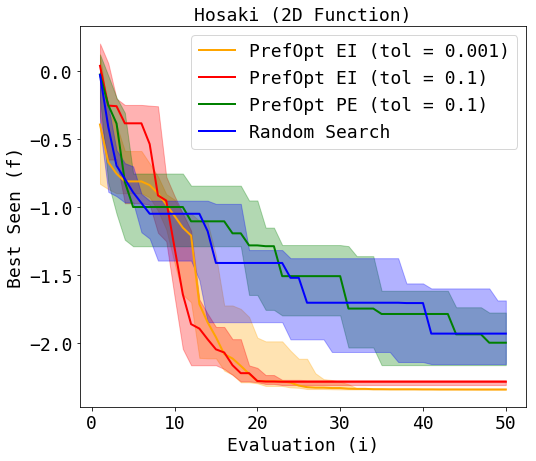} 
    \vspace{0.25mm}
  \end{minipage} 
  \begin{minipage}[b]{0.48\linewidth}
    \includegraphics[width=\linewidth]{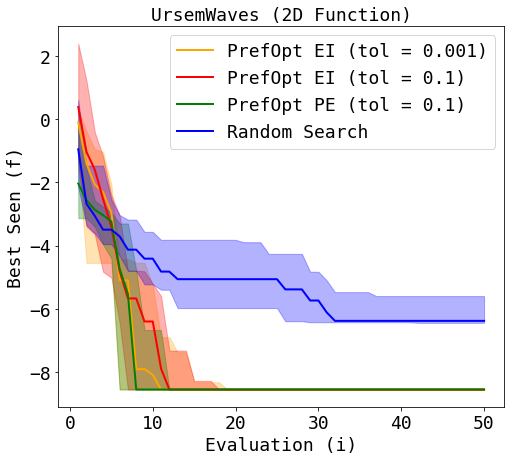}
    \vspace{0.25mm}
  \end{minipage}
  \begin{minipage}[b]{0.5\linewidth}
    \includegraphics[width=\linewidth]{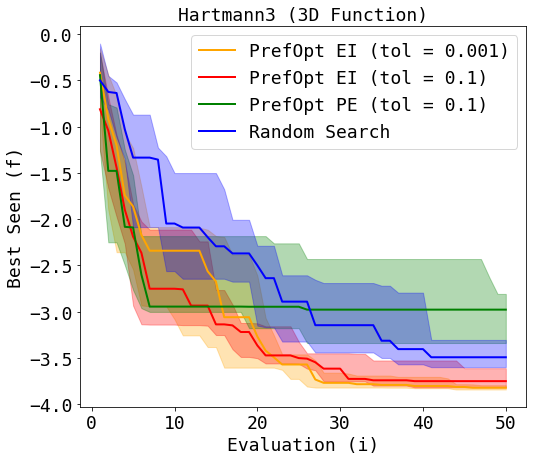} 
  \end{minipage}
  \hfill
  \begin{minipage}[b]{0.5\linewidth}
    \includegraphics[width=\linewidth]{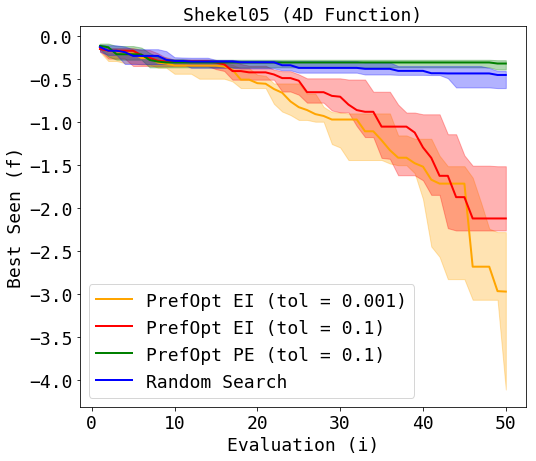} 
  \end{minipage}
        \vspace{-5mm}
  \caption{
      \label{fig:result_figures}
      Summary of PrefOpt optimization traces on a collection of synthetic test problems.
  }
\end{figure}

\vspace{-0.75mm}
To measure the effectiveness of our preference-based optimization method, we considered its efficiency in minimizing synthetic test functions
using only pairwise comparative observations.  At each iteration a query point ($\mathbf{x}^n$) is selected and compared against the current best point $(\mathbf{x}^b)$ using a 
test function ($f_\text{test}$).  The discrete preference observations were simulated in the following way : 
\vspace{-0.75mm}
\begin{align*}
\text{pref}(\mathbf{x}_1, \mathbf{x}_2) = \begin{cases}
\mathbf{x}_1 \approx \mathbf{x}_2 , \hspace{2mm} \text{if} \ \ |f_\text{test}(\mathbf{x}_1) - f_\text{test}(\mathbf{x}_2)| \leq \epsilon \\
\mathbf{x}_1 \succ \mathbf{x}_2 , \hspace{2mm} \text{else if} \ f_\text{test}(\mathbf{x}_1) < f_\text{test}(\mathbf{x}_2) \\
\mathbf{x}_1 \prec \mathbf{x}_2 , \hspace{2mm} \text{otherwise}
\end{cases} 
\end{align*}
\vspace{-0.75mm}
We initialized the search with  $2D+1$ samples from a latin hypercube sequence, where $D$ is the number of input parameters of the test function.
We repeated each experiment 10 times and reported the median and interquartile range of the best seen objective value after each iteration.  We included two settings of the 
tolerance parameter $\epsilon = \{0.001, 0.1\}$ and evaluated the expected improvement acquisition function, a ``pure exploration'' ($a_{\mathsf{PE}}$) acquisition function \cite{contal2013parallel, eric2008active}  
and random search \cite{bergstra2012random}.
\begin{align*}
a_{\mathsf{PE}}(\mathbf{x}^* ; \mathbf{z}  ) &= rbf(\mathbf{x}^*,\mathbf{x}^*, \pmb{\theta}) - \mathbf{k}_{*}^{\mathsf{T}} \mathbf{K}^{-1}\mathbf{k}_{*} \\
\mathbf{x}^n &= \argmax_{\mathbf{x}^*} \int a_{\mathsf{PE}}(\mathbf{x}^* ; \mathbf{z}  ) q(\mathbf{z} \ ; \ \pmb{\lambda} ) d\mathbf{z}
\end{align*}
We evaluated PrefOpt using several optimization test functions \cite{mccourt_test}. 
We observed the expected improvement acquisition function outperforming the baselines 
across the selected test functions \ref{fig:result_figures}.
Our approach seems work well under both settings of the tolerance parameter which is encouraging as it suggests our method might be useful even to users with less 
than expert ability to discern quality.
\vspace{-3mm}

\bibliographystyle{plain}
\bibliography{citations}

\end{document}